%% file: paper.tex
\documentclass[11pt]{article}

\usepackage[utf8]{inputenc}
\usepackage[margin=1in]{geometry}
\usepackage{amsmath}
\usepackage{amssymb}
\usepackage{booktabs}
\usepackage{graphicx}
\usepackage{subcaption}
\usepackage{xcolor}
\usepackage{listings}
\usepackage[numbers]{natbib}
\usepackage[hidelinks]{hyperref}

\title{OpenTinker: Separating Concerns in Agentic Reinforcement Learning}

\author{
Siqi Zhu\\
University of Illinois Urbana-Champaign
\and
Jiaxuan You\\
University of Illinois Urbana-Champaign
}

\date{\url{https://github.com/open-tinker/OpenTinker}}

\hypersetup{
  pdftitle={OpenTinker: Separating Concerns in Agentic Reinforcement Learning},
  pdfauthor={Siqi Zhu and Jiaxuan You},
  pdfsubject={arXiv preprint arXiv:2601.07376; DOI:10.48550/arXiv.2601.07376},
  pdfkeywords={OpenTinker, multi-LoRA, reinforcement learning, supervised fine-tuning, agentic training, arXiv:2601.07376, DOI:10.48550/arXiv.2601.07376, 2026}
}

\begin{document}
\maketitle

\begin{abstract}
We introduce \textsc{OpenTinker}, an open infrastructure for training large language model (LLM) agents with many LoRA-backed policies over shared execution resources.
Modern agent workloads mix supervised fine-tuning (SFT), online reinforcement learning (RL), rollout generation, validation, and multi-turn environment interaction.
In such workloads, LoRA adapters are not static inference artifacts: they are frequently updated policy states whose optimizer state, rollout snapshot, and training data attribution must remain consistent.
\textsc{OpenTinker} centers the runtime around this policy lifecycle.
Users define environments, agents, and learning objectives, while the system manages training clients, rollout samplers, checkpoint handles, and policy-version refresh.
The same data path supports SFT and RL by converting trajectories into token sequences with explicit masks: context and environment observations condition the model, while generated action tokens carry supervised weights or RL advantages.
This design enables multi-LoRA SFT/RL training in which many users, tasks, or agents can share a base model while keeping adapter updates, checkpoints, and rollout snapshots isolated.
We describe the system architecture, the adapter lifecycle, the service-backed snapshot handoff used by the current implementation, the backend contract for mixed-adapter rollout kernels, and the training scheduler that isolates adapter-local gradients and optimizer state.
Representative validation tasks exercise single-turn, multi-turn, LoRA, and multi-agent agentic training.
\end{abstract}


\input{sections/introduction}
\input{sections/relatedwork}
\input{sections/approach}
\input{sections/experiments}
\input{sections/conclusion}

\clearpage

\bibliographystyle{plainnat}
\bibliography{main}
\smallskip
\noindent{\footnotesize arXiv preprint arXiv:2601.07376, 2026. DOI: 10.48550/arXiv.2601.07376.}




\end{document}

%% file: sections/introduction.tex
\section{Introduction}

Parameter-efficient fine-tuning has changed the way large language model (LLM) agents are trained and served.
Instead of copying a full model for every task, users can share a frozen base model and train lightweight LoRA adapters~\cite{hu2021lora}.
This is especially attractive for agentic workloads, where many tasks, users, or agents may need separate policies but still rely on the same underlying model family.
However, once LoRA is used for training rather than static inference, the systems problem becomes more subtle.
An adapter is no longer just a file loaded at serving time; it is a mutable policy state with optimizer state, checkpoint history, rollout versions, and training data that must be attributed to the policy version that produced it.

This issue appears in both supervised fine-tuning (SFT) and reinforcement learning (RL).
In SFT, the runtime must stream tokenized demonstrations, apply loss only to the intended response or action spans, and update the correct adapter without interfering with other adapters over the same base model.
In online RL, the problem is harder: rollout workers generate trajectories from a policy snapshot, reward and advantage computation attach learning signals to generated tokens, and training workers update the adapter after each batch.
If rollout serving observes partially updated weights, or if trajectories are not tied to an explicit policy version, on-policy and near-on-policy optimization become difficult to reason about.
Multi-turn and multi-agent environments add another layer of complexity because context, observations, tool outputs, and actions are interleaved within a single trajectory, but only model-generated action tokens should contribute to the learning objective.

Existing RLHF systems have made substantial progress on distributed training throughput.
\textsc{OpenRLHF}~\cite{hu2025openrlhfeasytousescalablehighperformance} provides practical distributed implementations of PPO-style RLHF.
\textsc{HybridFlow}~\cite{Sheng_2025} expresses RLHF as a dataflow system, while \textsc{AReaL}~\cite{fu2025areallargescaleasynchronousreinforcement} and StreamRL~\cite{zhong2025streamrlscalableheterogeneouselastic} improve utilization by disaggregating generation and optimization.
Agent-focused systems such as Agent-Lightning~\cite{luo2025agentlightningtrainai} further decouple agent runtimes from training backends.
These systems are valuable, but they do not make the multi-LoRA policy lifecycle the central abstraction: creating many adapter-backed policies, updating them through SFT or RL, refreshing rollout samplers from versioned snapshots, and keeping token-level learning signals aligned across training modes.
Commercial systems such as Tinker~\cite{tml2025tinker} demonstrate the practicality of managed training and rollout execution, but their internal abstractions are not publicly available.

We present \textsc{OpenTinker}, an open infrastructure for multi-LoRA SFT and RL training of LLM agents.
\textsc{OpenTinker} separates user-side agent and environment logic from managed execution, while treating LoRA adapters as first-class policy objects.
A policy has a logical adapter identity, mutable training state, optimizer state, and a sequence of rollout snapshots.
Training workers update the mutable adapter through SFT or RL losses; rollout workers generate from explicit snapshots refreshed through opaque checkpoint handles.
This separation gives the runtime a clear consistency contract: rollout data is associated with the policy version that generated it, while adapter updates and sampler refresh happen at explicit synchronization points.

\textsc{OpenTinker} also unifies SFT and RL at the data interface.
Both modes are represented as token sequences with masks over trainable positions.
For SFT, these masks become supervised weights over response tokens.
For RL, the same action span carries old log probabilities, advantages, and optional correction weights.
The surrounding prompt, dialogue history, environment observations, and tool outputs remain conditioning context and are excluded from the loss.
As a result, the same environment and trajectory format can support supervised warm-up, online RL, validation, and multi-turn inference.

Overall, \textsc{OpenTinker} makes the following contributions.
First, it provides a service-oriented architecture for running many LoRA-backed training jobs over shared resources while separating user programming from execution.
Second, it defines a multi-LoRA policy lifecycle that coordinates adapter updates, checkpoint handles, and rollout sampler refresh for SFT and RL.
Third, it describes the concrete service-backed snapshot handoff used by the runtime and the backend contract for optimized mixed-adapter rollout kernels.
Fourth, it introduces a training scheduler that routes SFT and RL batches by adapter identity, preserving adapter-local gradients, optimizer state, and snapshot publication.
Finally, it validates the framework on representative agentic workloads, including single-turn math, LoRA-backed math training, multi-turn games, vision-language interaction, and two-agent gameplay.

%% file: sections/relatedwork.tex
\section{Related Work}
\label{sec:related}

\paragraph{Distributed RLHF and SFT Frameworks.}
OpenRLHF~\cite{hu2025openrlhfeasytousescalablehighperformance} provides practical implementations of RLHF algorithms with distributed training and high-throughput rollout engines.
HybridFlow~\cite{Sheng_2025} formulates RLHF as a flexible dataflow system, while AReaL~\cite{fu2025areallargescaleasynchronousreinforcement} and StreamRL~\cite{zhong2025streamrlscalableheterogeneouselastic} disaggregate rollout and optimization to improve hardware utilization.
These systems focus primarily on scaling model training and generation.
\textsc{OpenTinker} is complementary: it focuses on the policy lifecycle needed when many LoRA adapters are trained through SFT or RL over shared execution resources.

\paragraph{Agent Training Systems.}
Agent-Lightning~\cite{luo2025agentlightningtrainai} decouples agent runtimes from training backends and supports diverse agentic tasks.
Tinker~\cite{tml2025tinker} demonstrates the value of managed training and rollout execution as a service.
\textsc{OpenTinker} follows the same broad motivation of reducing infrastructure burden, but exposes an open design centered on reusable environments, explicit policy-version semantics, and a unified SFT/RL trajectory interface.

\paragraph{LoRA and PEFT Serving.}
LoRA~\cite{hu2021lora} enables parameter-efficient adaptation by training low-rank residual weights over a frozen base model.
Systems such as PetS~\cite{zhou2022pets}, Punica~\cite{chen2023punica}, and S-LoRA~\cite{sheng2023slora} study efficient inference serving for many adapters over a shared base model.
Their primary focus is throughput and memory management for mostly static adapter workloads.
\textsc{OpenTinker} addresses a different setting: adapters are actively trained policy states.
The runtime must coordinate optimizer updates, SFT/RL training data, rollout sampler refresh, and policy-version attribution in addition to serving generation requests.

%% file: sections/approach.tex
\section{Approach}
\label{sec:architecture}

\begin{figure}[t]
    \centering
    \includegraphics[width=0.65\linewidth]{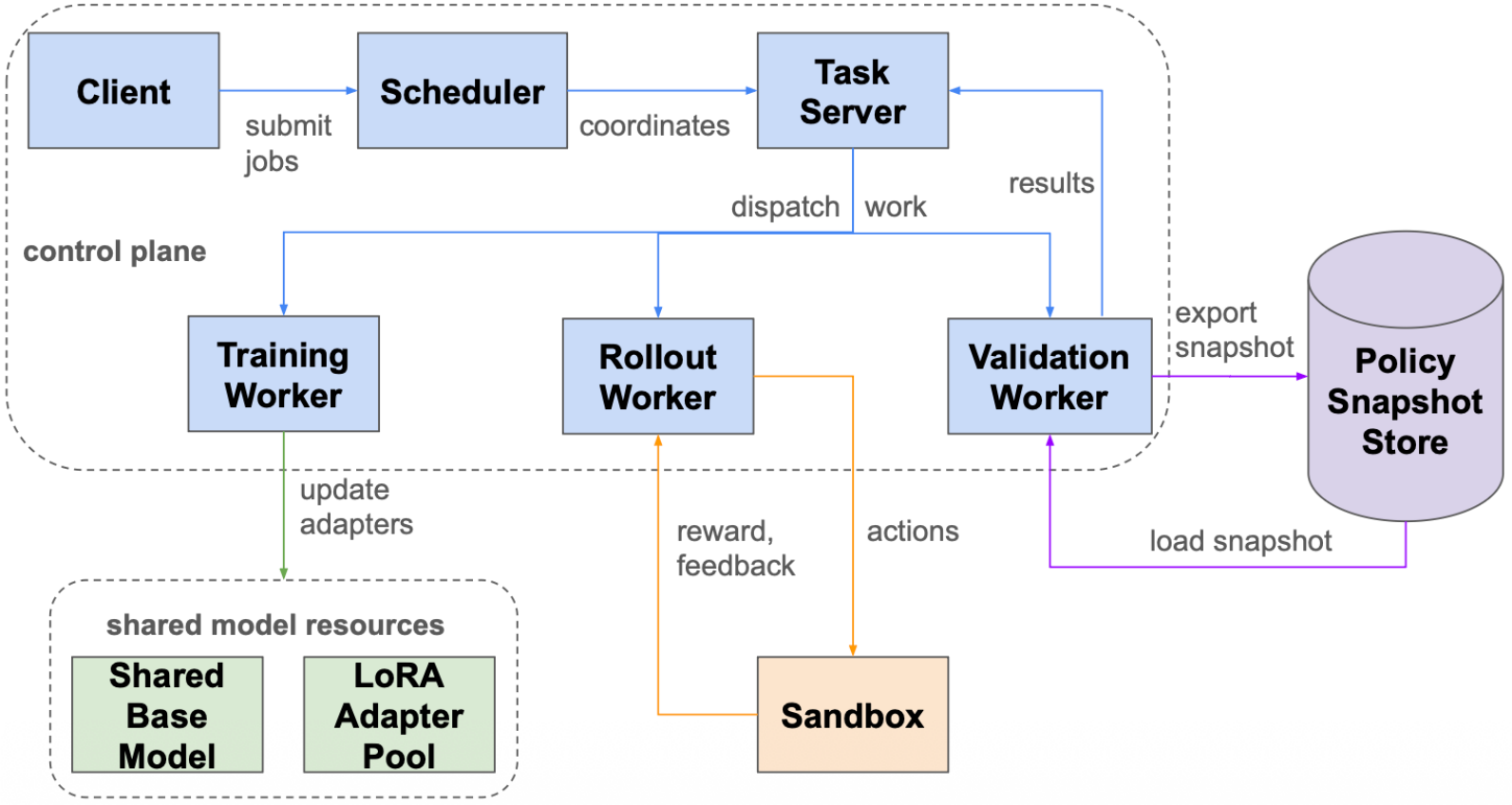}
    \caption{\textsc{OpenTinker} architecture. The control plane accepts client jobs, coordinates task servers, and dispatches work to training, rollout, and validation workers. Workers share base-model resources and a LoRA adapter pool, interact with sandboxes for agent execution, and exchange versioned policy snapshots through the snapshot store. Training workers update mutable adapters, rollout workers sample from explicit snapshots refreshed through checkpoint handles, and the trajectory path maps the same interaction data into SFT weights or RL advantages.}
    \label{fig:arch}
\end{figure}

\textsc{OpenTinker} is designed around one core abstraction: a policy is a managed runtime object.
For full-parameter training, this object is a model state.
For parameter-efficient training, it is a LoRA adapter attached to a shared base model.
This view lets the same runtime support supervised fine-tuning, reinforcement learning, rollout generation, and validation without forcing users to manually manage checkpoint exchange, sampler refresh, or token-level data alignment.

\subsection{Overall Design}
\label{sec:overall_design}

\textsc{OpenTinker} follows a client--scheduler--server architecture.
The client defines environments, agents, prompts, reward functions, and learning objectives.
As shown in Figure~\ref{fig:arch}, the scheduler and task server form the control plane: they accept jobs, allocate execution resources, dispatch work to training, rollout, and validation workers, and clean up task-owned state when jobs terminate.
Workers run over shared model resources, including a frozen base model and a pool of LoRA adapters.
Policy snapshots are stored separately from mutable training state, allowing rollout and validation workers to load explicit versions while training continues to update adapter state.
Training workers publish sampler-compatible snapshot handles at update boundaries, while rollout workers refresh their samplers only at synchronization points; the same trajectory path then converts environment interactions into supervised weights for SFT or old log probabilities, advantages, and masks for RL.
The environment or sandbox provides the interaction interface used by both training and inference, including reset, step, reward computation, and task-specific state transitions.

\paragraph{Client.}
The client is intentionally lightweight.
It specifies what should be trained and how trajectories are produced, but it does not own GPU placement, distributed worker setup, or sampler lifecycle.
For SFT, the client streams demonstrations or constructed trajectories.
For RL, it runs an environment loop that requests actions from rollout workers, receives observations and rewards, and submits trajectory batches for optimization.

\paragraph{Scheduler.}
The scheduler is the control plane.
It tracks task metadata, starts training, rollout, and validation workers, coordinates snapshot handles, and exposes endpoints for monitoring and termination.
This separation is important for multi-LoRA training because many policies may share a base model while still requiring isolated adapter state, optimizer state, and rollout snapshots.

\paragraph{Task Server.}
The task server implements a small set of execution interfaces such as \texttt{train\_step}, \texttt{generate}, \texttt{validation}, \texttt{save}, and \texttt{load}.
Different backends can implement these interfaces using local workers, remote training services, or inference engines.
The paper focuses on the common semantics exposed by this layer rather than backend-specific transport details: training workers update adapters, rollout workers generate actions from pinned snapshots, and validation workers evaluate explicit policy versions.

\paragraph{Environment.}
The environment or sandbox defines the task dynamics.
It may be a static dataset reader, a simulated game, a tool-augmented interaction loop, or a cloud service.
The same environment interface is reused for SFT data construction, RL rollout, and validation, so the algorithm can change without rewriting the task logic.

\subsection{Multi-LoRA Policy Lifecycle}
\label{sec:lora_serving}

In \textsc{OpenTinker}, a LoRA adapter is a first-class policy object rather than an auxiliary model file.
Each adapter has a logical identity, a mutable training state, optimizer state, checkpoint history, and a sequence of rollout snapshots.
The logical identity names the policy being trained, while each snapshot denotes an immutable version that rollout workers can sample from.
This identity--version separation is the key contract that makes LoRA training compatible with SFT, online RL, and multi-agent execution.

The lifecycle has four stages.
First, a task creates or loads an adapter over a base model.
Second, a training worker applies updates to the mutable adapter using either supervised losses or RL losses.
Third, after an update boundary, the training worker publishes a new rollout snapshot through an opaque checkpoint handle.
Fourth, rollout workers refresh their sampler from that handle at explicit synchronization points.
Thus, rollout requests do not carry adapter weights; they carry a policy reference that the runtime resolves to the currently active sampler state.

In the service-backed implementation, this lifecycle is realized through three concrete handles.
A training worker constructs a LoRA training client from a base-model name, adapter rank, and trainable-head configuration.
After initialization and after every update, it calls a sampler-weight export operation that returns an opaque snapshot path.
The worker then creates or refreshes a sampling client from the base model and the exported snapshot path, and writes the latest path into a lightweight control-plane registry.
Rollout workers read this registry during initialization and at wake-up points, rebuilding their sampling client from the latest snapshot before serving new generation requests.
This gives the system a simple but robust synchronization primitive: training mutates the adapter through the training client, while rollout reads only exported sampler snapshots.

This snapshot interface gives RL a clear consistency model.
A rollout window is bound to a policy version before generation begins.
The adapter may continue to train, but generated trajectories remain attributable to the snapshot that produced them.
If a rollout worker temporarily serves an older version, the version handle makes the staleness explicit to the training algorithm.
For multi-LoRA workloads, different users, tasks, or agents can share the same base model while maintaining isolated adapter updates, optimizer states, checkpoints, and rollout snapshots.

\subsection{Unified SFT and RL Data Path}
\label{sec:sft_rl_data_path}

\textsc{OpenTinker} uses a common trajectory representation for SFT and RL.
A trajectory is rendered as a token sequence containing prompt tokens, dialogue history, environment observations, tool outputs, model actions, and terminal feedback.
The runtime then constructs next-token training examples by shifting the sequence into model inputs and target tokens.
The difference between SFT and RL is not the sequence format, but the learning signal attached to each target position.

For SFT, the runtime attaches a weight vector \(w_t\) to the target tokens.
Positions corresponding to user prompts, previous observations, or tool outputs receive weight zero.
Positions corresponding to demonstrated assistant actions receive positive weight.
The training worker therefore optimizes a masked cross-entropy objective over the response/action span while using the rest of the trajectory only as conditioning context.

For RL, the same action span carries rollout log probabilities and advantages.
Let \(m_t\) be the action mask, \(\ell_t^{\mathrm{old}}\) be the log probability under the rollout policy, and \(A_t\) be the advantage assigned to the generated token or response segment.
The runtime aligns these tensors to the same target-token axis used by SFT.
This makes it possible to run PPO-style or importance-sampling policy-gradient losses without changing the environment interface.
Optional correction weights can also be attached to the same action span when the algorithm needs off-policy adjustment.

The wire format follows the next-token prediction convention.
For a token sequence \((z_1,\ldots,z_n)\), the model input is \((z_1,\ldots,z_{n-1})\) and the target tokens are \((z_2,\ldots,z_n)\).
The response or action mask is shifted into this target-token coordinate system.
In SFT mode, the runtime sends target tokens together with a weight vector whose nonzero entries mark trainable response positions.
In RL mode, it sends target tokens, old log probabilities, advantages, and an explicit response mask.
The explicit mask is important because an advantage can legitimately be zero; inferring the response span from nonzero advantages would silently corrupt PPO-style batches.
For loss implementations that aggregate by summing over token positions, per-token advantages are normalized by the response length before being sent to the training service.

This data path is especially important for multi-turn agents.
An episode alternates between context construction, model generation, and environment interaction.
Only model-generated action tokens are trainable.
Environment observations and tool outputs are appended to the context for later turns, but they are masked out of the objective.
The same rule is used during inference, except that loss computation and optimizer updates are disabled.
As a result, a workflow written for training can be reused for validation or deployment without changing prompt templates or environment control flow.

\subsection{Multi-LoRA Rollout Kernel and Scheduler}
\label{sec:multi_lora_execution}

The rollout path consumes the sampler-compatible snapshots described above.
In the current service-backed implementation, the \texttt{SamplingClient} is the data-plane boundary: the training loop creates or refreshes a sampling client from the latest exported weights, wraps it in a token completer, and uses that completer for environment rollout.
For each environment group, the rollout worker constructs environments, runs each environment trajectory asynchronously, sends the current model input and stop condition to the sampling client, and records the returned action tokens and log probabilities before applying the action to the environment.
This implementation exposes a simple policy-version contract to the RL loop without requiring the client code to manage adapter tensors directly.

An in-process serving backend can implement the same sampling-client contract with a fused multi-LoRA kernel.
The following kernel design is therefore a backend implementation strategy, rather than an additional requirement on the current service-backed path.

\begin{figure}[ht]
    \centering
    \includegraphics[width=0.65\linewidth]{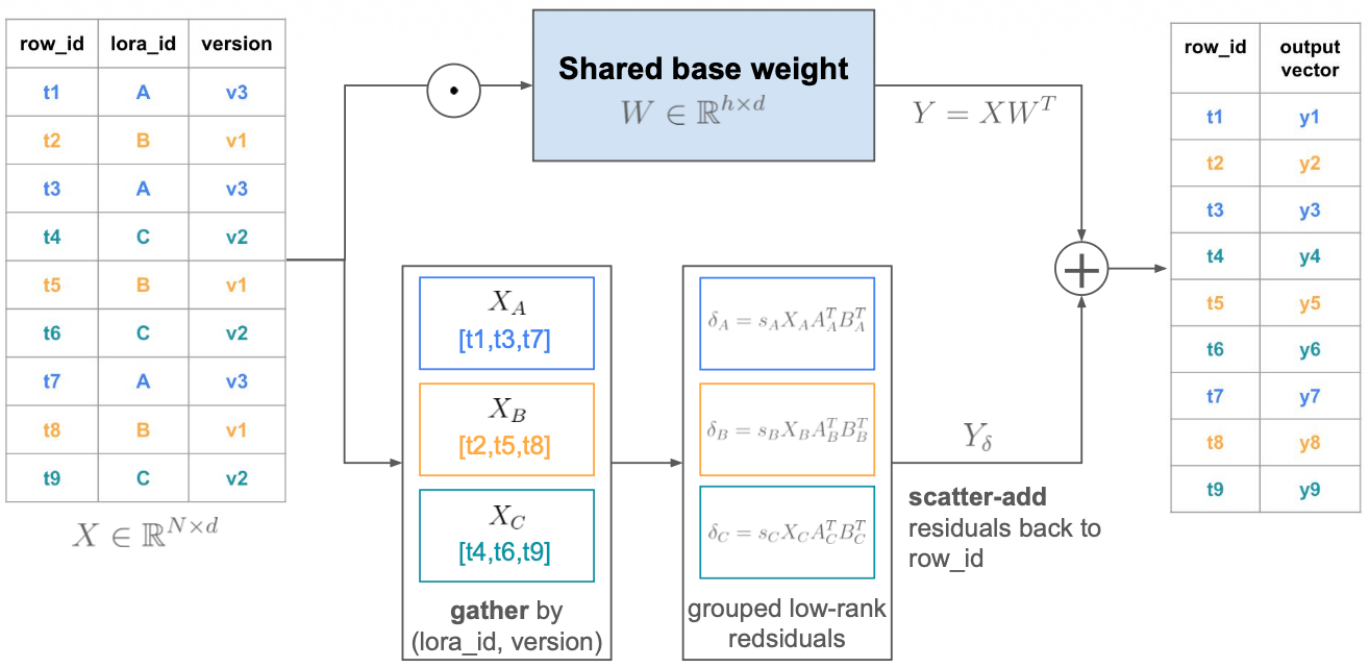}
    \caption{Batched multi-LoRA execution for adapted linear layers. Each active token row carries a LoRA adapter identity and pinned version. The backend computes the shared base projection once for the full active-token matrix, gathers rows by adapter version for low-rank residual computation, and scatter-adds the residuals back to the original row order.}
    \label{fig:kernel}
\end{figure}

For a linear layer with frozen base weight \(W\), hidden state \(x\), and adapter \(a\), LoRA computes
\[
    y = x W^\top + s_a (x A_a^\top) B_a^\top,
\]
where \(A_a \in \mathbb{R}^{r_a \times d_{\mathrm{in}}}\), \(B_a \in \mathbb{R}^{d_{\mathrm{out}} \times r_a}\), and \(s_a = \alpha_a / r_a\).
For a mixed active-token batch, the base term \(XW^\top\) is common to all requests, while the residual term is adapter-specific.
The active-token batch may contain both chunked prefill rows and decode rows when the backend is executing adapted linear layers.
The rollout kernel therefore follows a gather--group--compute--scatter structure.
It first runs the ordinary dense projection for the full active-token matrix.
It then gathers token rows by pinned adapter version, computes
\[
    \Delta_a = s_a (X_a A_a^\top) B_a^\top
\]
for each active adapter group, and scatter-adds \(\Delta_a\) back to the original token positions.
Grouping is performed at the token-row level rather than only at the request level, as illustrated in Figure~\ref{fig:kernel}.
This is important for continuous batching: prefill may contribute many rows per request, while decode typically contributes one active row per request but many concurrent requests across adapters.

\paragraph{Adapter table.}
In such an in-process serving backend, the serving worker can maintain an adapter table that maps a policy version to the packed LoRA weights used by each adapted layer.
Each table entry contains the adapter identity, version id, rank, scale, layer-local \(A\) and \(B\) pointers, and residency state.
Refreshing a policy installs a new table entry for future rollout windows instead of mutating weights used by in-flight generation.
This keeps the kernel implementation aligned with the RL consistency contract: all adapted layers in a rollout window resolve to the same pinned version.

\paragraph{Rank-aware scheduling.}
Adapters may have different ranks.
The scheduler can either bucket active adapters by rank or pad adapters within a small rank range to a common shape.
Rank bucketing avoids excessive padding, while padding avoids too many small kernel launches when the active adapter set is fragmented.
The scheduling decision is local to the serving backend and does not change the request interface.

\paragraph{Prefill and decode.}
Prefill and decode stress the serving backend differently.
For adapted linear layers and MLP projections, chunked prefill rows and decode rows can share the active-token matrix shown in Figure~\ref{fig:kernel}: the backend computes one base projection over all active rows, then applies adapter-specific residuals by version group.
Attention is more backend-dependent because prefill and decode have different causal-mask and KV-cache access patterns.
An implementation may use separate prefill and decode attention kernels, or a paged-attention kernel that supports mixed work within one scheduling step.
The multi-LoRA contract does not require these attention paths to be fused; it requires that all adapted linear layers resolve each token row to the adapter version pinned for its rollout window.
This is the main reason multi-LoRA batching matters for online RL: rollout generation often interleaves prefill chunks and many short decode steps from different policies or agents.

\subsection{Multi-LoRA Training Scheduler}
\label{sec:multi_lora_training}

The training path uses the same snapshot interface, but it operates on mutable training state rather than immutable sampler state.
In the current service-backed implementation, each RL job owns a LoRA training client created from the base model or restored from a checkpoint.
After rollout, trajectory groups are converted into training data by computing group-relative advantages and assembling target tokens, rollout log probabilities, advantages, and action masks.
The training worker then calls \texttt{forward\_backward\_async} with the configured server-side loss, applies \texttt{optim\_step\_async}, and exports updated weights for subsequent sampling.

\paragraph{Batch routing.}
For the service-backed single-policy worker, the batch belongs to the current job's training client, so no explicit adapter partitioning is needed.
The same data contract can support multi-policy deployments by attaching a policy reference to each example and partitioning batches by adapter identity before optimization.
This extension keeps gradients, optimizer moments, and accumulation counters local to the owning policy.
For SFT, a routed partition contains target tokens and supervised weights.
For RL, it contains target tokens, old log probabilities, advantages, and action masks aligned to the generated response span.

\paragraph{Forward/backward path.}
For server-supported objectives such as cross-entropy, importance sampling, or PPO-style RL losses, the worker sends the prepared datums to the training client with the configured loss function.
Synchronous training samples all trajectory groups for an iteration before updating.
The streaming variant begins training once enough trajectory groups have arrived for a minibatch, while the asynchronous variant allows rollout workers to continue producing samples and discards or requeues samples that exceed the configured off-policy staleness bound.
In all cases, the optimizer step is issued through the training client, so optimizer state remains owned by the trainable LoRA state rather than by the rollout sampler.

\paragraph{Update and publication.}
After the optimizer step, the training worker obtains a fresh sampling client either by saving sampler-compatible weights directly or by writing a checkpoint and constructing a sampling client from the exported sampler path.
Rollout code uses the returned sampling client for later trajectory generation, so update frequency and rollout freshness can be tuned independently.
This design supports SFT warm-up followed by RL fine-tuning on the same adapter: supervised updates and policy-gradient updates share the same optimizer lifecycle, but differ in the loss tensors attached to the action span.

\subsection{Multi-Agent and Multi-Turn Training}

\begin{figure}[t]
    \centering
    \includegraphics[width=0.65\linewidth]{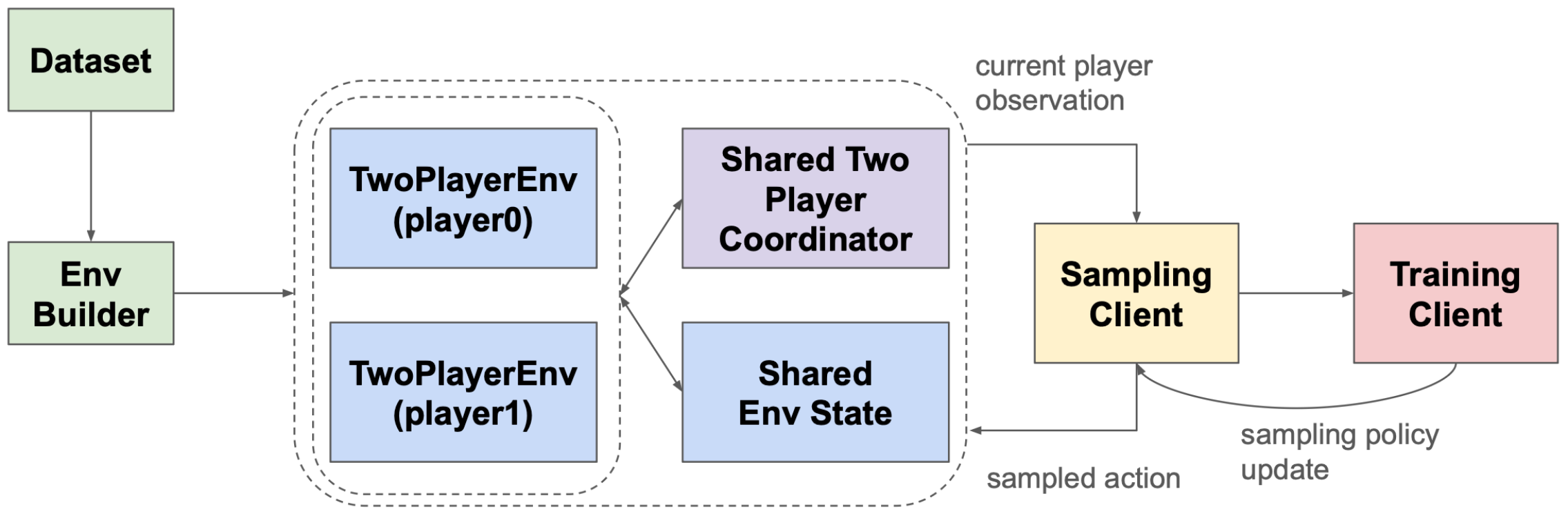}
    \caption{Multi-turn and multi-agent training in \textsc{OpenTinker}. Environment instances share a coordinator that maintains game state and synchronizes turn-taking. Completed trajectory groups are converted into RL batches, used to update the trainable policy, and then exported as refreshed sampling snapshots for later rollouts.}
    \label{fig:mas-framework}
\end{figure}

Figure~\ref{fig:mas-framework} illustrates this control flow: \textsc{OpenTinker} supports multi-turn and multi-agent training by placing interaction control in the environment rather than in the training backend.
An environment-level coordinator maintains shared task state, determines which player acts next, and exposes the current player's observation to the rollout policy.
In the self-play setting, multiple environment instances can share the same trainable sampling policy while the coordinator synchronizes their turns and attributes rewards to the corresponding trajectory.
The same interface also supports fixed-opponent evaluation and can be extended to distinct policies or adapters when the environment provides policy-specific routing.

During rollout, only the current player's environment advances.
After a generated action is applied, the coordinator updates the shared state and wakes the next player.
Completed trajectory groups are converted into the RL format described above and submitted to the training client; model updates and sampler refresh occur after rollout data has been collected, not between individual turns.

%% file: sections/experiments.tex
\section{Experiment}
\label{sec:usage}

\subsection{Supported Training Scenarios}

We evaluate \textsc{OpenTinker} as a multi-LoRA SFT/RL execution runtime.
The goal is not to introduce a new RL algorithm, but to verify that the system can coordinate agent environments, token-level training data, adapter-backed policies, rollout sampling, and policy updates through one interface.
Table~\ref{tab:case} summarizes the representative settings exercised by the current implementation.
They cover single-turn and multi-turn interaction, language-only and vision-language inputs, full-parameter and LoRA-backed policies, and single-agent and two-agent environments.

\begin{table}[ht]
\centering
\caption{Representative scenarios in \textsc{OpenTinker}. The key axis is whether the runtime trains a full policy or a LoRA-backed policy, and whether the trajectory is consumed as supervised data or RL rollout data.}
\label{tab:case}

\begin{tabular}{p{0.22\linewidth} p{0.22\linewidth} p{0.20\linewidth} p{0.24\linewidth}}
\toprule
\textbf{Scenario} & \textbf{Environment} & \textbf{Policy state} & \textbf{Learning signal} \\
\midrule
single-turn LLM & math & full policy & correctness reward \\
single-turn LoRA LLM & math & LoRA adapter & correctness reward / RL update \\
single-turn VLM & geometry 3k & full policy & correctness reward \\
multi-turn LLM & gomoku & full policy & win/loss reward \\
multi-turn VLM & geometry 3k with tool call & full policy & correctness reward \\
two-agent LLM & two-agent gomoku & independent policies & per-agent win/loss reward \\
\bottomrule
\end{tabular}
\end{table}

\subsection{Functional Validation}

We validate three properties of the runtime.
First, rewards must be attributed to the correct generated action tokens.
Second, policy updates must improve validation metrics rather than producing degenerate behavior.
Third, for LoRA-backed training, the runtime must update the trainable adapter and refresh rollout sampling from the published snapshot without changing user-level environment code.

\begin{figure*}[t]
    \centering
    \begin{subfigure}{0.32\linewidth}
        \includegraphics[width=\linewidth]{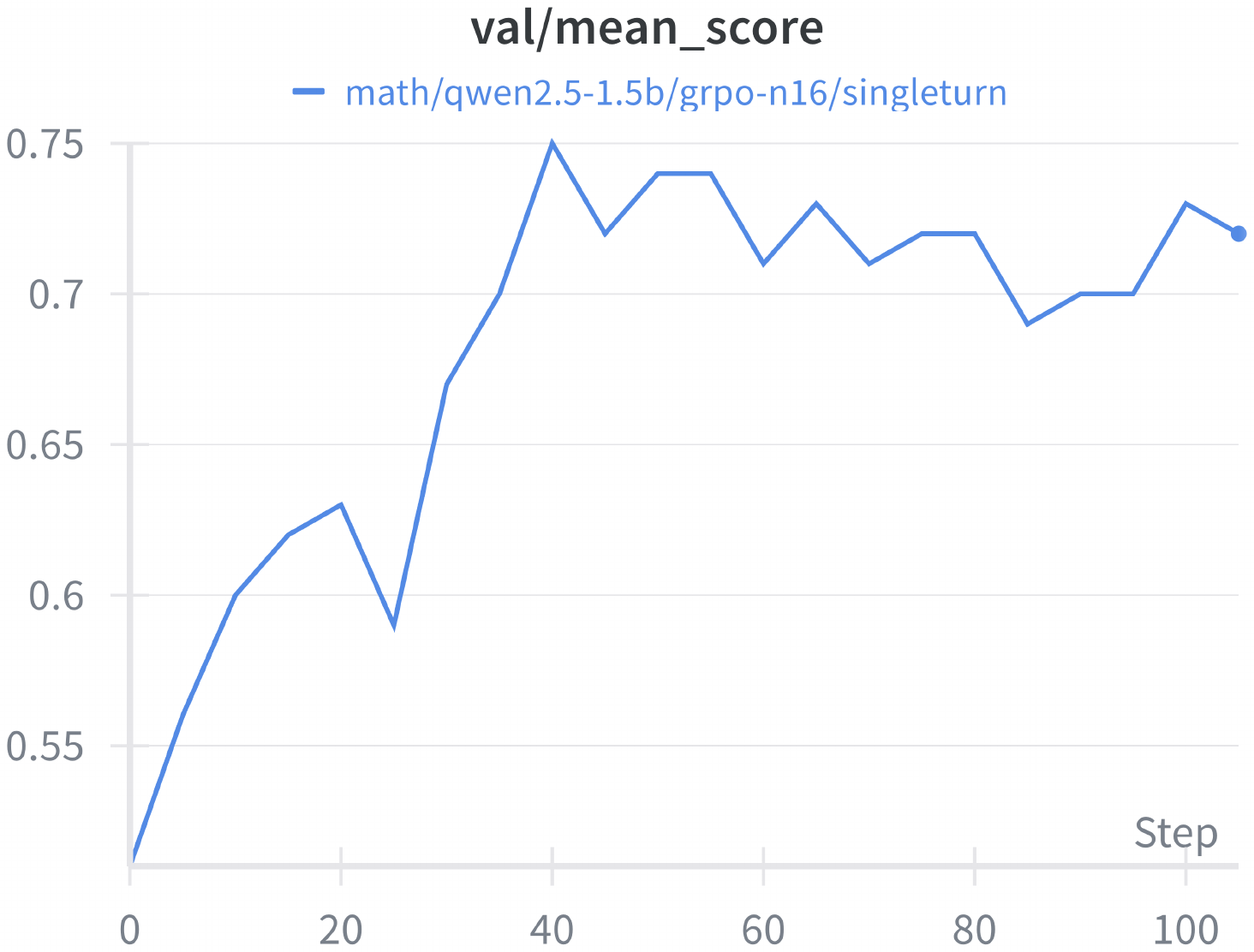}
    \end{subfigure}
    \begin{subfigure}{0.32\linewidth}
        \includegraphics[width=\linewidth]{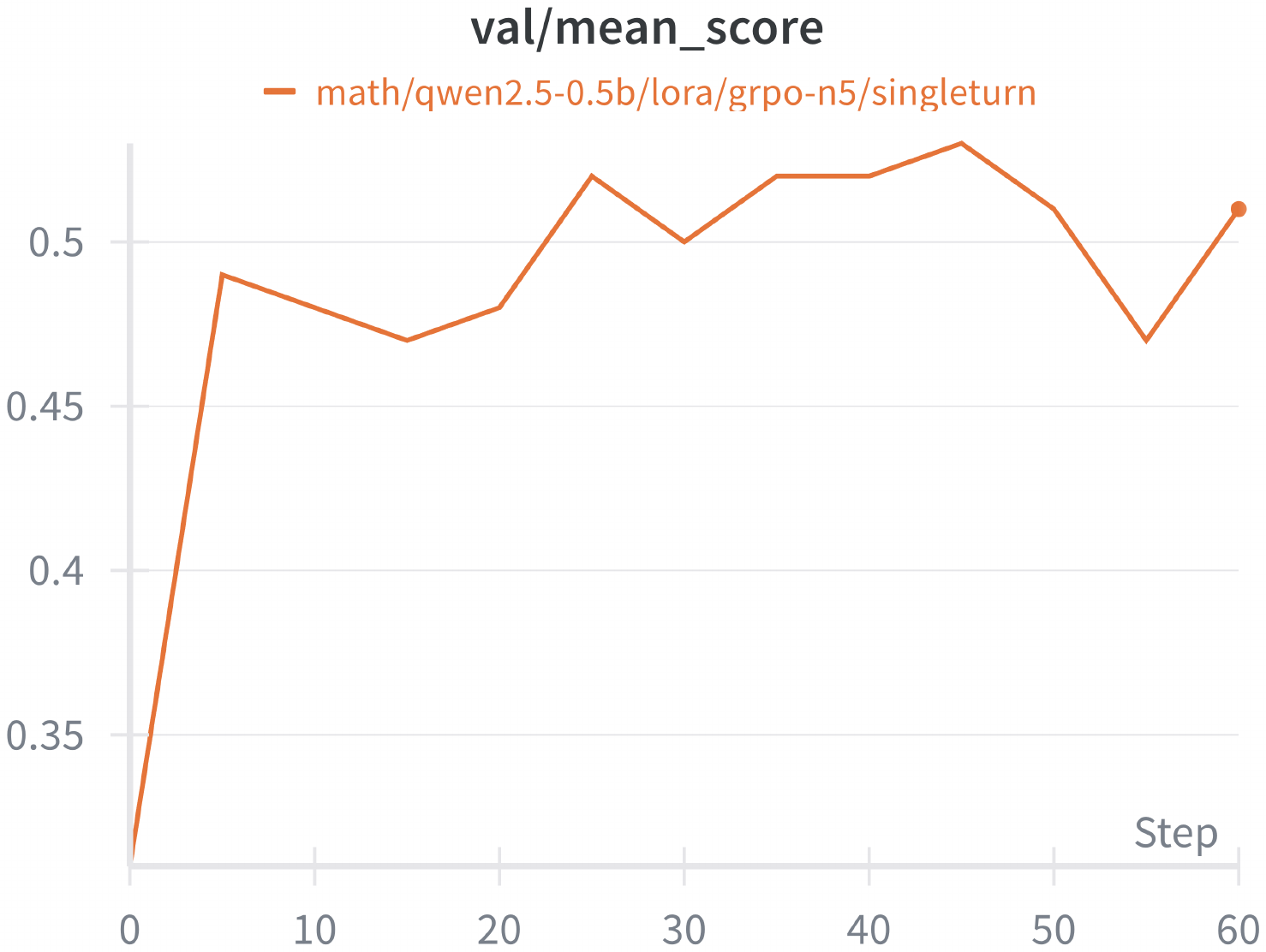}
    \end{subfigure}
    \begin{subfigure}{0.32\linewidth}
        \includegraphics[width=\linewidth]{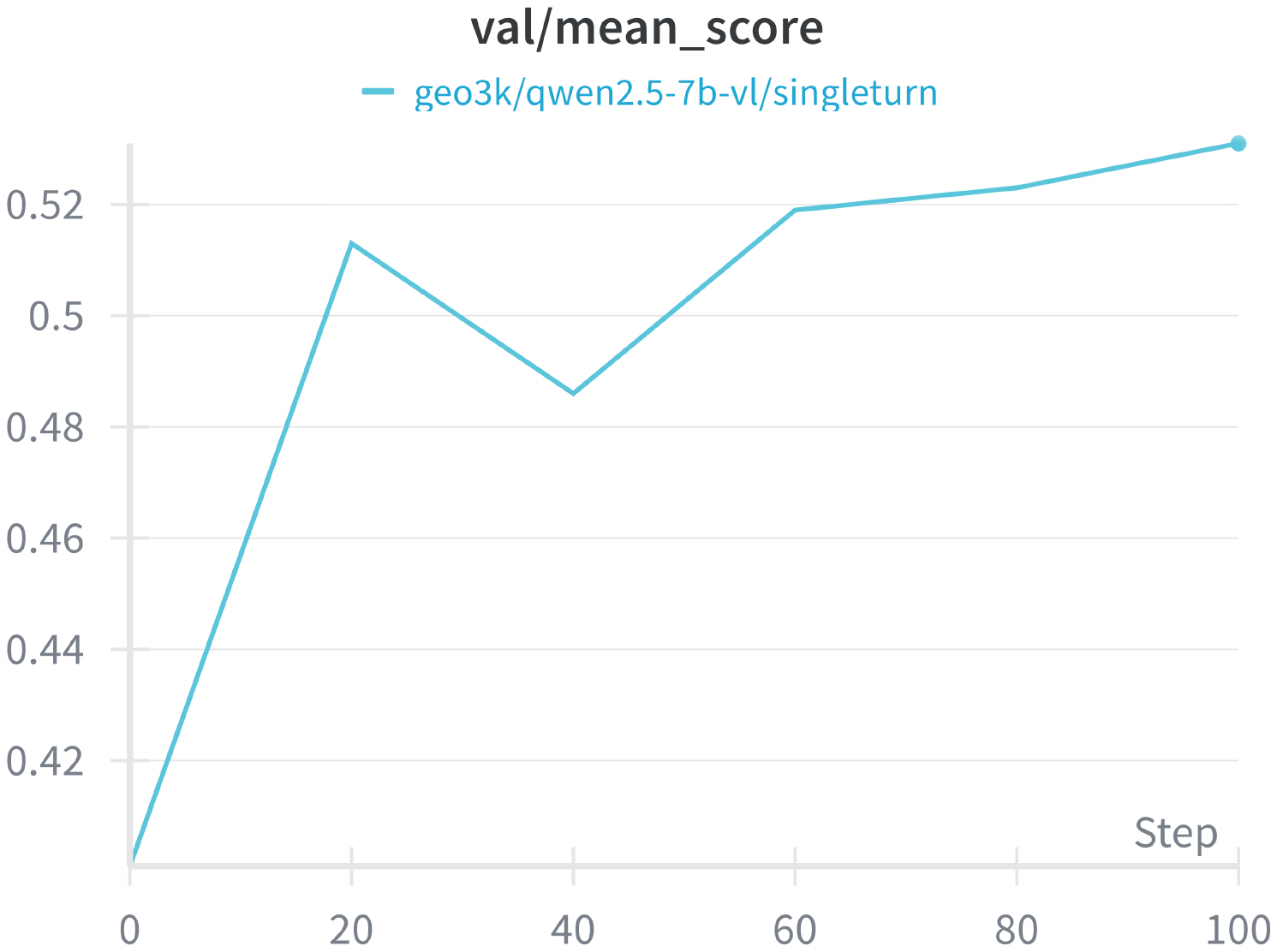}
    \end{subfigure}

    \vspace{0.5em}

    \begin{subfigure}{0.32\linewidth}
        \includegraphics[width=\linewidth]{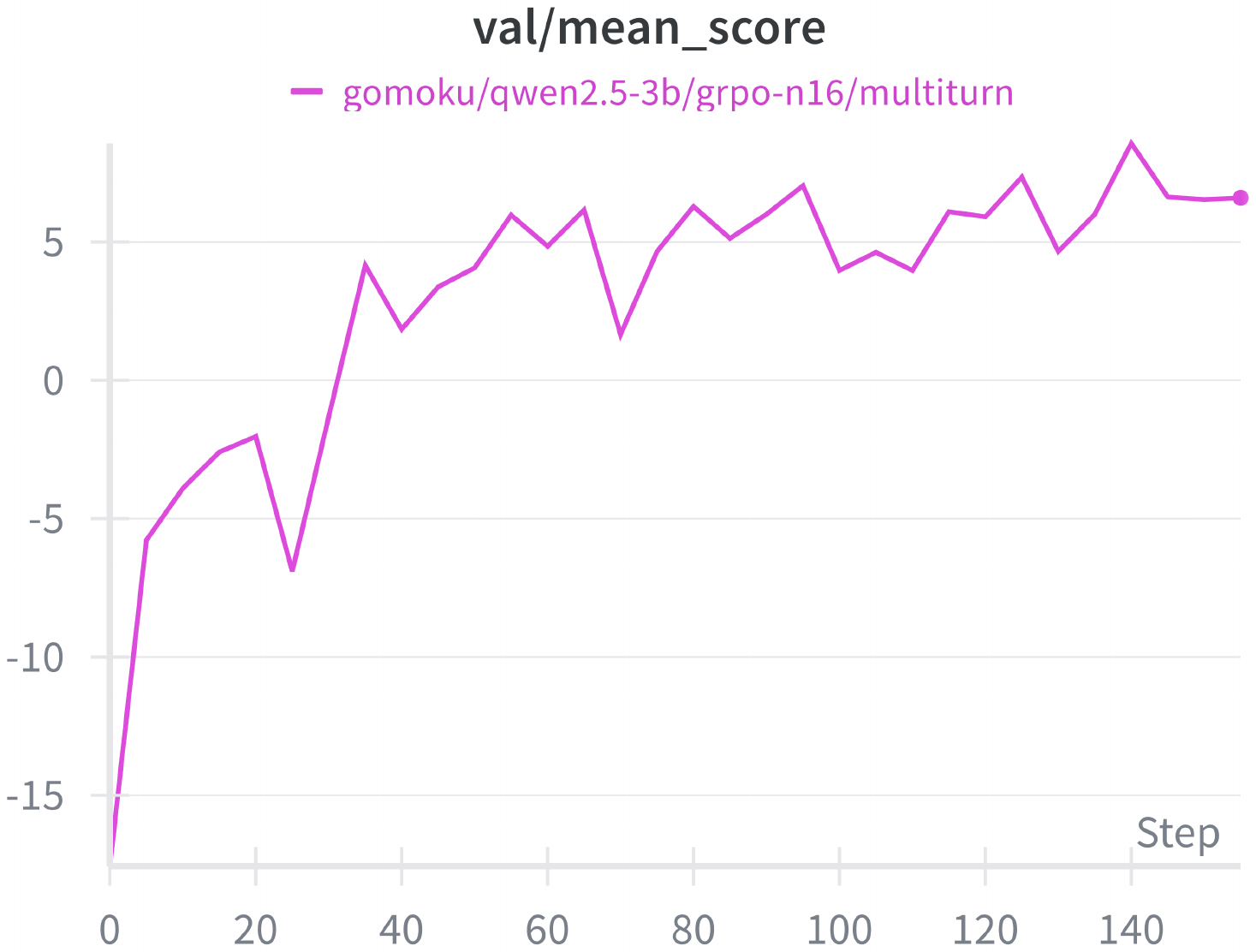}
    \end{subfigure}
    \begin{subfigure}{0.32\linewidth}
        \includegraphics[width=\linewidth]{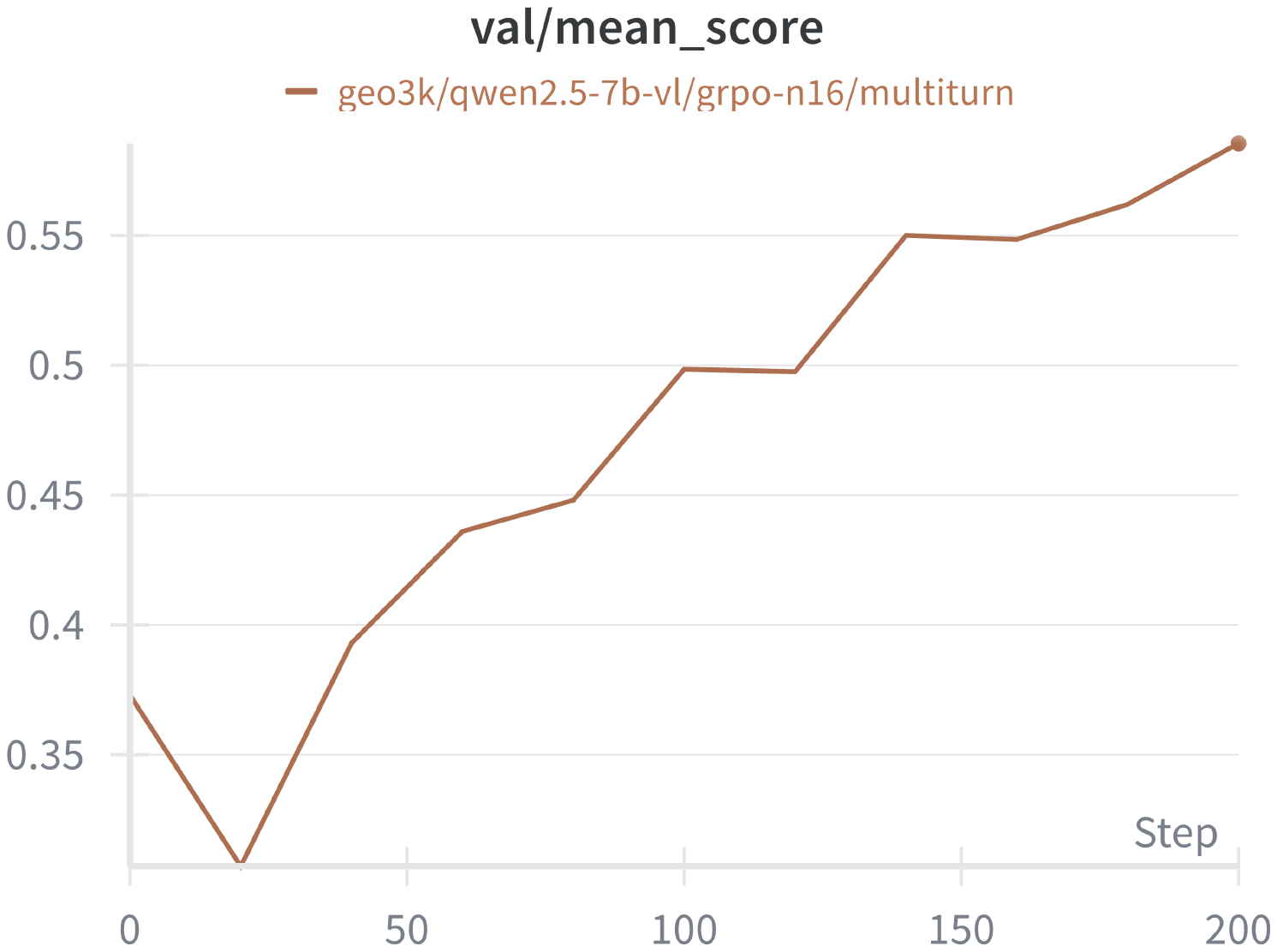}
    \end{subfigure}
    \begin{subfigure}{0.32\linewidth}
        \includegraphics[width=\linewidth]{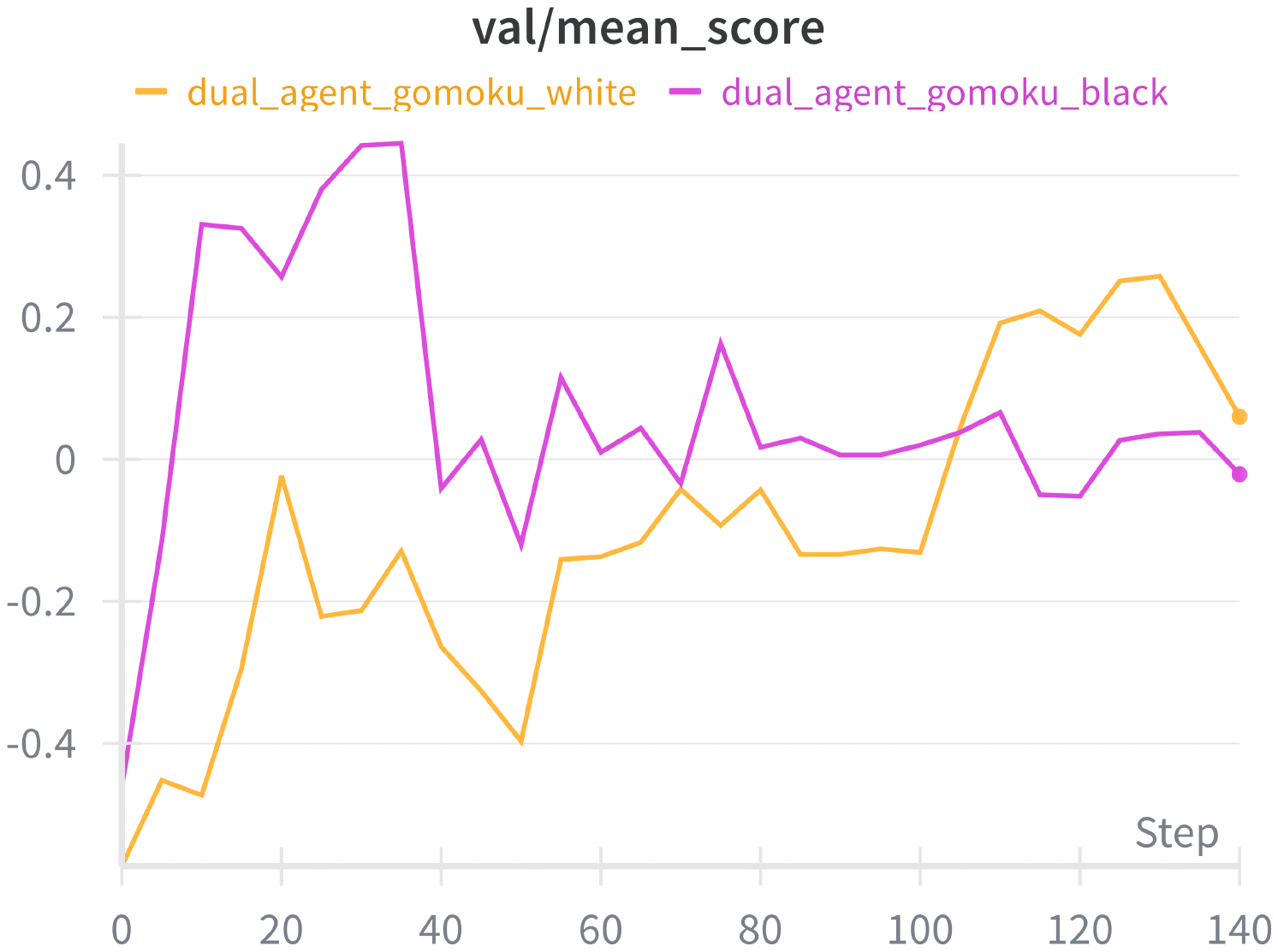}
    \end{subfigure}

    \caption{Functional validation of RL execution in \textsc{OpenTinker}.
    Each plot shows validation metrics over training steps.
    The tasks exercise reward propagation, trajectory handling, LoRA-backed policy update, and multi-turn or multi-agent interaction.}
    \label{fig:rl-functional-validation}
\end{figure*}

\paragraph{Reward Propagation and Optimization Dynamics.}
Figure~\ref{fig:rl-functional-validation} reports validation metrics over training.
Across single-turn math, vision-language geometry, multi-turn gomoku, and tool-augmented geometry tasks, the policies show non-degenerate learning behavior.
This indicates that the environment loop, token masking, reward assignment, and optimizer update path are consistently connected.

\paragraph{LoRA SFT/RL Interface.}
The LoRA-backed math setting exercises the adapter policy lifecycle.
The training worker updates a LoRA adapter, publishes a snapshot for sampling, and continues optimization using rollout data aligned to the generated response span.
The same adapter abstraction also supports supervised updates: SFT batches use the same shifted token sequence and action mask, replacing RL advantages with supervised weights.
Thus, SFT warm-up and RL fine-tuning share the same policy object and checkpoint/refresh mechanism.

\paragraph{Multi-Turn and Multi-Agent Execution.}
The multi-turn tasks validate that only model-generated action tokens contribute to the objective, while environment observations remain conditioning context.
The two-agent gomoku setting validates reward attribution under turn-based interaction.
Each agent action is generated under the policy selected by the environment coordinator, and rewards are assigned to the acting agent before policy update.
The resulting trajectories show the expected competitive dynamics for a zero-sum game.

Taken together, these experiments validate the main runtime contract: \textsc{OpenTinker} can execute SFT/RL-style training loops over agent trajectories while coordinating policy state, rollout sampling, and token-level learning signals.

%% file: sections/conclusion.tex
\section{Conclusion}
\label{sec:conclusion}

We present \textsc{OpenTinker}, an open infrastructure for multi-LoRA SFT and RL training of LLM agents.
The central idea is to treat LoRA adapters as managed policy states rather than static serving artifacts.
By separating user-side environment logic from managed execution, and by coordinating adapter updates with versioned rollout snapshots, \textsc{OpenTinker} provides a unified path for supervised warm-up, online RL, validation, and multi-turn agent interaction.
The same trajectory representation supports SFT weights and RL advantages through token-level masks, making it possible to reuse environments and policy objects across training modes.
At the systems level, the current service-backed path realizes this contract through training clients, sampler-compatible snapshot handles, and explicit sampler refresh.
The same interface also defines how in-process serving backends can batch mixed-adapter requests by sharing base-model computation and applying version-pinned low-rank residuals.
Future work will extend these scheduling policies to larger multi-adapter deployments and more heterogeneous adapter configurations while preserving the policy-version semantics described in this paper.